\documentclass[letterpaper, 10 pt, conference]{ieeeconf}  

\IEEEoverridecommandlockouts                              

\usepackage[english]{babel}
\usepackage{stfloats}
\usepackage{graphicx}
\usepackage{epstopdf}
\usepackage{subfigure}
\usepackage{xcolor}

\usepackage{amsthm}

\graphicspath{{figures/}}
\usepackage{makecell}
\usepackage{sidecap}
\usepackage{amsmath}
\usepackage{multirow}
\usepackage{amssymb}
\usepackage{amsmath}
\usepackage{bm}
\newsavebox\CBox

\usepackage{amsthm,amsmath,amssymb}
\usepackage{mathrsfs}

\usepackage{verbatim}

\usepackage{hyperref}

\usepackage{graphicx}
\usepackage{subfigure}
\usepackage{siunitx}
\usepackage{underscore}

\usepackage[ruled]{algorithm2e}

\usepackage{epstopdf}
\usepackage{booktabs}
\usepackage{cite,soul}

\usepackage{titlesec}

\usepackage{soul} 

\titleformat{\subsubsection}[runin]{\bfseries}{}{}{}

\titlespacing{\subsubsection}{0pt}{\parskip}{-\parskip}

\title{\LARGE \bf
Chance-Aware Lane Change with High-Level Model Predictive Control Through Curriculum Reinforcement Learning
}

\author{Yubin Wang, Yulin Li, Zengqi Peng, Hakim Ghazzai, 
and Jun Ma 
\thanks{This work was supported in part by the National Natural Science Foundation of China under Grant 62303390; and in part by the Project of Hetao Shenzhen-Hong Kong Science and Technology Innovation Cooperation Zone under Grant HZQB-KCZYB-2020083. \textit{(Corresponding Author: Jun Ma.)}}
\thanks{Yubin Wang, Yulin Li, and Zengqi Peng are with the Robotics and Autonomous Systems Thrust, The Hong Kong University of Science and Technology (Guangzhou), Guangzhou, China (email: ywang575@connect.hkust-gz.edu.cn; yline@connect.ust.hk; zpeng940@connect.hkust-gz.edu.cn) }
\thanks{Hakim Ghazzai is with the Computer, Electrical and Mathematical Science and Engineering Division, King Abdullah University of Science and Technology, Thuwal, Saudi Arabia (email: Hakim.Ghazzai@kaust.edu.sa)}
\thanks{Jun Ma is with the Robotics and Autonomous Systems Thrust, The Hong Kong University of Science and Technology (Guangzhou), Guangzhou, China, also with the Division of Emerging Interdisciplinary Areas, The Hong Kong University of Science and Technology, Hong Kong SAR, China, and also with the HKUST Shenzhen-Hong Kong Collaborative Innovation Research Institute, Futian, Shenzhen, China (e-mail: jun.ma@ust.hk).}}

\begin{document}

\maketitle
\thispagestyle{empty}
\pagestyle{empty}

\begin{abstract}
Lane change in dense traffic typically requires the recognition of an appropriate opportunity for maneuvers, which remains a challenging problem in self-driving. In this work, we propose a chance-aware lane-change strategy with high-level model predictive control (MPC) through curriculum reinforcement learning (CRL). In our proposed framework, full-state references and regulatory factors concerning the relative importance of each cost term in the embodied MPC are generated by a neural policy.
Furthermore, effective curricula are designed and integrated into an episodic reinforcement learning (RL) framework with policy transfer and enhancement, to improve the convergence speed and ensure a high-quality policy.
The proposed framework is deployed and evaluated in numerical simulations of dense and dynamic traffic. 
It is noteworthy that, given a narrow chance, 
the proposed approach generates high-quality lane-change maneuvers such that the vehicle merges into the traffic flow with a high success rate of $96\%$. Finally, our framework is validated in the high-fidelity simulator under dense traffic, demonstrating satisfactory practicality and generalizability.

\end{abstract}

\section{Introduction}

Chance-aware lane change is geared towards maneuvering the ego vehicle to track and occupy the recognized narrow and moving chance with an appropriate safe margin, such that the ego vehicle merges into the traffic flow. 
However, it is still an open and challenging problem on how to generate a satisfactory trajectory and feasible lane-change maneuvers. 
First, explicit prediction in terms of motions and states of traffic flow is demanded to infer the exact pose for the execution of lane change.  
Also, the time window needs to be identified dynamically for lane-change maneuvers during the planning horizon. 
To address these challenges, it is imperative to develop a motion planner for chance-aware lane change with strong adaptiveness towards such dense and dynamic traffic. 

As one of the widely used optimization-based approaches, model predictive control (MPC) has gained wide popularity for its effectiveness in optimizing the trajectory while dealing with various constraints for self-driving~\cite{mukai2017model,dixit2019trajectory, eiras2021two}.  
However, in complex scenarios such as the aforementioned chance-aware lane-change problem, it is nearly impossible to properly set all the handcrafted factors of MPC, e.g., time-varying constraints corresponding to the dynamic identification of time windows for lane-change maneuvers.  
On the other hand, reinforcement learning (RL) has shown promising results in generating effective and agile maneuvers for self-driving by learning a driving strategy through trial-and-error~\cite{kiran2021deep, nishitani2020deep, fuchs2021super, song2021autonomous}. 
However, it is a common problem that pure RL-based methods suffer from instability and limited safety guarantees of control policies.

Learning-based MPC stands as a promising hybrid paradigm for devising control strategies, wherein critical elements of MPC can be acquired through classical machine learning models or deep neural networks. 
A typical routine for parameterizing the MPC with crucial factors relies on learning the system model \cite{kabzan2019learning,hewing2019cautious}.
Also, the formulation of constraints within the MPC scheme can benefit from deep learning techniques~\cite{bae2020cooperation,bae2022lane}. 
On the other hand, the integration of MPC and RL has been explored for motion planning tasks due to
its outstanding flexibility and adaptiveness to challenging problems.
As one of the seminal works in this area, a Gaussian distribution is harnessed to model the high-level policy in \cite{song2020learning, song2022policy}, with which the traversal time is defined as a sequence of decision variables, such that the MPC can be parameterized. 
Hence, the quadrotor accomplishes the task of flying through a swinging gate. However, the state references of MPC are fixed as the gate states, which could degrade the performance of agile flight. 
Additionally, in \cite{wang2023learning}, 
SE(3) decision variables modeled by deep neural networks, are further designed as the references of MPC.
\textcolor{black}{Essentially, this approach manifests the effectiveness in traversing a moving and rotating gate.} 
Nevertheless, the weight modulation under the MPC scheme is empirical, which hinders the generalization of the method to dense and dynamic environments with higher complexity. 

In this paper, a novel learning-based MPC framework for chance-aware lane-change tasks is proposed, where the augmented decision variables are designed to parameterize the MPC. Specifically, {we make use of a neural policy to learn} full states as references of MPC and also their regulatory factors which can automatically determine the relative importance of references within the planning horizon. 
We incorporate curriculum reinforcement learning (CRL) with policy transfer and enhancement to learn the optimal policy progressively with ordered curricula.
The contributions of this paper are listed as follows:

(1) We propose a novel learning-based MPC framework that incorporates full-state references and regulatory factors which can modulate the relative importance of each cost term within the cost functions. 
This facilitates the effective extraction and adjustment of all crucial information for optimizing the solution quality of MPC, leading to improved adaptiveness to dense and dynamic traffic.
    
(2) To improve the policy quality and avoid unstable learning, we present CRL with policy transfer and enhancement to learn a neural policy, which achieves faster convergence and higher reward compared to other baselines.

(3) The proposed approach is validated through numerical simulations under dense and dynamic traffic, where improved safety and effectiveness are demonstrated through comparative experiments. Furthermore, the practicality and generalizability are illustrated through experimental validations in the high-fidelity simulator.

\section{Problem Statement}
\label{sec:statement}

\subsection{Vehicle Dynamic Model}
 
The bicycle model in \cite{ge2021numerically} is used,
where the state vector of the vehicle is defined as 
$\mathbf{x}=\left[\begin{array}{llllll}
    p_x & p_y & \varphi & v_x & v_y & \omega
    \end{array}\right]^{\top}$,
where $p_x$ and $p_y$ denote the X-coordinate and Y-coordinate position of the vehicle’s center of mass, $\varphi$ is the heading angle, $v_x$ and $v_y$ are the longitudinal and lateral speed, and $\omega$ represents the yaw angular velocity. Also, we integrate the actions into a vector as $\mathbf{u}=\left[\begin{array}{ll}a & \delta\end{array}\right]^{\top}$, where $a$ and $\delta$ are the acceleration and steering angle. 
Then, the nonlinear dynamic model $f_{\mathrm{dyn}}$ of the vehicle in discrete time is given by:
\begin{equation}
    \begin{aligned}
        \mathbf{x}_{t+1} 
        &= \mathbf{x}_{t} + f_{\mathrm{dyn}}( \mathbf{x}_{t},  \mathbf{u}_{t}) d_t \\
        &= \left[\begin{array}{c}
    p_{x,t} + (v_{x,t} \cos \varphi_t -v_{y,t} \sin \varphi_t)d_t \\
    p_{y,t} + (v_{x,t} \sin \varphi_t +v_{y,t} \cos \varphi_t)d_t \\
    \varphi_t + \omega_t d_t\\
    v_{x,t} + a_t d_t\\
    \frac{L_k \omega_t d_t - k_f \delta_t v_{x,t} d_t - m v^2_{x,t} \omega_t d_t}{m v_{x,t} - (k_f + k_r) d_t}\\
    \frac{I_z v_{x,t}\omega_t + L_k v_{y,t} d_t - l_f k_f \delta_t v_{x,t} d_t}{I_z v_{x,t} - (l^2_f k_f + l^2_r k_r) d_t }
    \end{array}\right],
    \end{aligned}
    \label{dyn}
    \end{equation}
where $t$ represents the current time, $d_t$ denotes the step time, $m$ is the mass of the vehicle, $l_f$ and $l_r$ are the distance from the center of mass to the front and rear axle, $k_f$ and $k_r$ are the cornering
stiffness of the front and rear wheels, $L_k = l_f k_f - l_r k_r$, and $I_z$ is the polar moment of inertia.

\begin{figure}[!t] 
    \centering   
    \includegraphics[trim=0 0 0 0, width=0.9\linewidth]{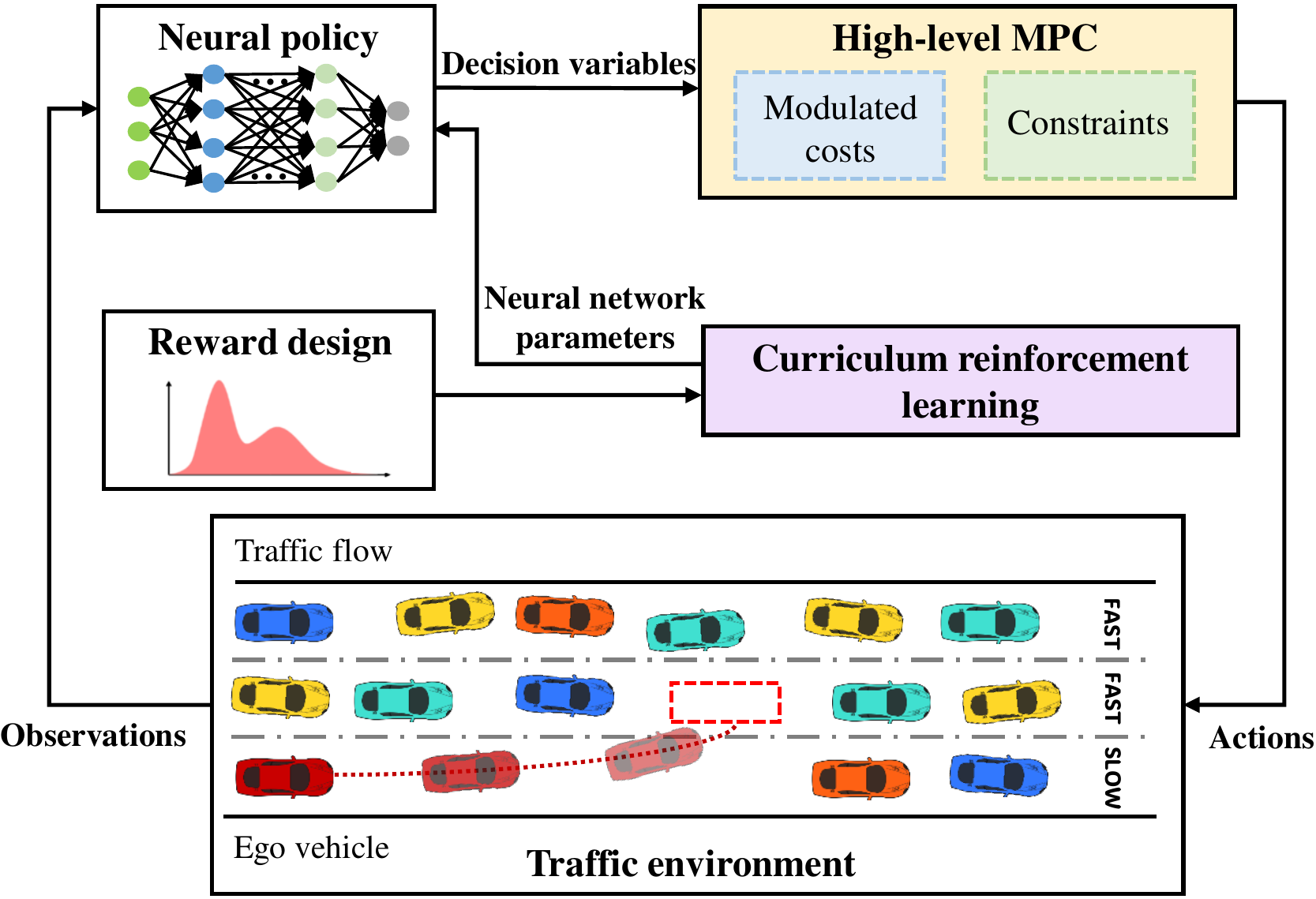}
   \caption{Overview of our proposed framework for chance-aware lane-change problems. The recognized dynamic chance for lane change is visually represented through the use of a red dashed rectangle. 
   }
    \label{frame}
\end{figure}

\subsection{Chance-Aware Lane-Change Task}
In this work, the focus lies on the spatial-temporal characteristics of the entirety of traffic flow, as opposed to individual surrounding vehicles. 
Consequently, the driving behaviors of each vehicle within the traffic flow are unrestricted, encompassing motions such as accelerating, braking, and steering. 
Notably, to preserve the underlying topology of the modeled traffic flow, lane-change behaviors of vehicles within the traffic flow are prohibited. 
To facilitate the modeling of traffic flow, we assume that the traffic flow maintains consistent speed with inherent uncertainties. 
Additionally, the ego vehicle is obstructed by a dead-end, where the front vehicles drive slowly. The dynamic chance for the ego vehicle to perform lane change is recognized as the gap of the traffic flow on the middle lane. The illustration of the traffic is depicted in Fig.~\ref{frame}. 

In this context, the objective of the chance-aware lane-change mission is to plan the optimal state trajectory $\mathbf{x}_k^*, \forall k \in\{0, 1, \cdots, N\}$ towards the goal states $\mathbf{x}_g$ and obtain a sequence of optimal control commands $\mathbf{u}_k^*, \forall k \in\{0, 1, \cdots, N-1\}$ over a receding horizon $N$, such that the ego vehicle merges into the traffic flow successfully on the target lane. Concurrently, the lane change necessitates the adoption of appropriate maneuvers by the ego vehicle within a specified temporal window, which occurs precisely before any potential collision with the front vehicles.

\section{High-Level Model Predictive Control with Augmented Decision Variables}
\label{sec:mpc}

\subsection{MPC Formulation}

To solve the chance-aware lane-change problem, we formulate a nonlinear MPC with augmented decision variables over the receding horizon $N$:

\begin{align}
&\min _{\mathbf{x}_{0: N}, \mathbf{u}_{0: N-1}} J_{x_{N}} + \sum_{k=0}^{N-1}\left(J_{x_k}+ J_{u_k}+J_{\Delta u_k}+J_{\operatorname{tra}, k}\right) \nonumber \\ 
& \quad \quad \quad \quad = \boldsymbol{\delta}_{x, N}^T \mathbf{Q}_{x} \boldsymbol{\delta}_{x, N}+\sum_{k=0}^{N-1} (
\boldsymbol{\delta}_{x, k}^T \mathbf{Q}_{x} \boldsymbol{\delta}_{x, k}+ 
\boldsymbol{\delta}_{u, k}^T \mathbf{Q}_{u} \boldsymbol{\delta}_{u, k} \nonumber\\ 
& \quad \quad \quad \quad +\Delta_{u, k}^T \mathbf{Q}_{\Delta_u} \Delta_{u, k}
+\boldsymbol{\delta}_{\mathrm{tra}, k}^T \mathbf{Q}_{\mathrm{tra}}\left(t_{\mathrm{tra}}, k\right) \boldsymbol{\delta}_{\mathrm{tra}, k}) \nonumber\\
&\text{\qquad s.t.} \quad \mathbf{x}_{k+1}=\mathbf{x}_k+f_{\mathrm{dyn}}\left(\mathbf{x}_k, \mathbf{u}_k\right) d_t, \nonumber\\
& \quad \quad \quad \quad \mathbf{x}_0=\mathbf{x}_{\text {init }}, \mathbf{u}_{-1}=\mathbf{u}_{\text {init }}, \nonumber\\
& \quad \quad \quad \quad p_{y, \operatorname{min}} \leq p_{y} \leq p_{y, \operatorname{max}}, \nonumber\\
& \quad \quad \quad \quad a_{\min } \leq a \leq a_{\max }, \nonumber\\ 
& \quad \quad \quad \quad -\delta_{\max } \leq \delta \leq \delta_{\max}, \label{MPC}
\end{align}
where $\boldsymbol{\delta}_{x, k}=\left(\mathbf{x}_k -\mathbf{x}_g\right)$ denotes the difference between the current states $\mathbf{x}_k$ and the goal states $\mathbf{x}_g$, $\boldsymbol{\delta}_{\mathrm{tra}, k} = \left(\mathbf{x}_k -\mathbf{x}_\mathrm{tra}\right)$ represents the difference between
the current states $\mathbf{x}_k$ and the learnable full-state references $\mathbf{x}_\mathrm{tra}$, 
$\boldsymbol{\delta}_{u, k}=\mathbf{u}_k$ is a regularization term for control commands $\mathbf{u}_k$,
$\boldsymbol{\Delta}_{u, k}=\left(\mathbf{u}_k-\mathbf{u}_{k-1}\right)$ is another regularization term considering the variation of control commands for driving comfort. The quadratic cost terms in (\ref{MPC}) are weighted by diagonal matrices $\mathbf{Q}_{x}$, $\mathbf{Q}_{\mathrm{tra}}\left(t_{\mathrm{tra}}, k\right)$, $\mathbf{Q}_{u}$, and $\mathbf{Q}_{\Delta_u}$. 
The initial states and initial control commands are represented by $\mathbf{x}_{\text {init}}$ and $\mathbf{u}_{\text {init}}$, respectively. 
The Y-coordinate position is bounded within $p_{y, \operatorname{min}}$ and $p_{y, \operatorname{max}}$ introduced by the road width. Additionally, control commands are constrained by $a_{\min}$, $a_{\max}$, $-\delta_{\max}$, and $\delta_{\max}$ considering the physical limits of vehicle dynamics.

\subsection{MPC Parameterization with Augmented Decision Variables} 
We introduce the learnable full-state references $\mathbf{x}_\mathrm{tra}$ as the desired pose and speed of a \textcolor{black}{lane-change maneuver}:
\begin{equation}
 \mathbf{x}_\mathrm{tra} = \left[p_{x, \operatorname{tra}}, p_{y, \operatorname{tra}}, \varphi_{\operatorname{tra}}, v_{x, \operatorname{tra}}, v_{y, \operatorname{tra}}, \omega_{\operatorname{tra}} \right]^\top, 
\end{equation}
where $\mathbf{x}_\mathrm{tra}$ are intermediate states for temporary tracking of MPC. 
To endow the vehicle with the ability to automatically balance the importance between tracking $\mathbf{x}_\mathrm{tra}$ and tracking $\mathbf{x}_g$, the weighting matrix $\mathbf{Q}_{\mathrm{tra}}\left(t_{\mathrm{tra}}, k\right)$ with adaptive adjustment is defined as:
\begin{equation}
\mathbf{Q}_{\mathrm{tra}}\left(t_{\mathrm{tra}}, k\right)=\mathbf{Q}_{\max }  \exp \left(-\gamma  \left(k  d_t-t_{\mathrm{tra}}\right)^2\right),    
\end{equation}
where
$\mathbf{Q}_{\max}$ is a learnable maximum of $\mathbf{Q}_{\mathrm{tra}}$, $\gamma \in \mathbb{R}_{+}$ is the exponential decay rate for costs in terms of tracking $\mathbf{x}_\mathrm{tra}$, and $t_{\mathrm{tra}}$ is a learnable tracking time reference, which determines the opportune timing for a lane change. 
In order to further modulate the relative importance of each state reference respectively, $\mathbf{Q}_{\max}$ is defined as: 
 \begin{equation}
\begin{aligned}
    &\mathbf{Q}_{\max } =\operatorname{diag}(Q_{p_{x,\operatorname{tra}}}, Q_{p_{y,\operatorname{tra}}}, Q_{\varphi_{\operatorname{tra}}}, Q_{v_{x,\operatorname{tra}}},  \\
    &Q_{v_{y,\operatorname{tra}}}, Q_{\omega_{\operatorname{tra}}}).
\end{aligned}
\end{equation}
Specifically, $Q_{p_{x,\operatorname{tra}}}$ and $Q_{p_{y,\operatorname{tra}}}$ in $\mathbf{Q}_{\max}$ are assigned relatively large values whereas $\mathbf{x}_\mathrm{tra}$ and $t_{\mathrm{tra}}$ tend to confine themselves to small values. The considerable discrepancy in magnitude could potentially hinder the training process. Therefore, $\mathbf{Q}_{\max}$ adopts the proportion form of $\mathbf{Q}_{x}$, i.e., $\tilde{\mathbf{Q}}_{\max}
\leftarrow \mathbf{Q}_{\max} \odot \mathbf{Q}_{x}$, where $\odot$ is the Hadamard product for  element-wise multiplication. 

Therefore, the regulatory factors $\mathbf{Q}_{\max}$ and $t_{\mathrm{tra}}$  modulate the cost functions of MPC collaboratively, which balance the importance between tracking $\mathbf{x}_\mathrm{tra}$ and tracking $\mathbf{x}_g$. Therefore, the \textcolor{black}{lane-change} behavior is divided into two distinct phases. During the initial phase, the ego vehicle prepares for the maneuvers of the lane-change behavior by tracking $\mathbf{x}_\mathrm{tra}$; subsequently, in the latter phase, the ego vehicle concentrates on tracking $\mathbf{x}_g$ to effectively merge into the traffic flow.

We integrate all decision variables to an augmented decision vector $\mathbf{z}$ as:
\begin{equation}
\begin{aligned}
    &\mathbf{z}=[p_{x, \operatorname{tra}}, p_{y, \operatorname{tra}}, \varphi_{\operatorname{tra}}, v_{x, \operatorname{tra}}, v_{y, \operatorname{tra}}, \omega_{\operatorname{tra}}, Q_{p_{x,\operatorname{tra}}}, \\
    &Q_{p_{y,\operatorname{tra}}}, Q_{\varphi_{\operatorname{tra}}} , Q_{v_{x,\operatorname{tra}}}, Q_{v_{y,\operatorname{tra}}}, Q_{\omega_{\operatorname{tra}}}, t_{\operatorname{tra}}]^\top \in \mathbb{R}^{13}.
\end{aligned}
\end{equation}

In this sense, the MPC is parameterized by \textcolor{black}{the} decision vector $\mathbf{z}$. By feeding MPC with different $\mathbf{z}$, different corresponding optimal state trajectories are generated, denoted as $\boldsymbol{\xi}^*(\mathbf{z}) =f_\mathrm{MPC}(\mathbf{z})$ (i.e., $\boldsymbol{\xi}^*(\mathbf{z})=\left\{\mathbf{x}_k^*(\mathbf{z})\right\}_{k=0}^N$). $f_\mathrm{MPC}$ is defined as the mapping function of MPC.
Hence, we can incorporate the episodic RL technique for policy search \cite{song2020learning, song2022policy, wang2023learning} to determine the optimal policy 
$\pi^*$
that automatically tune the augmented decision variables.

\section{Curriculum Reinforcement Learning with Policy Transfer and Enhancement}
\label{sec:rl}

\subsection{Observation and Policy Representation}

The observation vector of the ego vehicle is defined as follows: 
    \begin{equation}
    \begin{aligned}
    & \mathbf{o} = [
    p_{x,\operatorname{init}}, p_{y,\operatorname{init}}, \varphi_{\operatorname{init}}, v_{x,\operatorname{init}}, 
    p_{x,\operatorname{init}}^{\operatorname{c}}, p_{y,\operatorname{init}}^{\operatorname{c}}, v_{x,\operatorname{init}}^{\operatorname{c}}, \\
    & p_{x,\operatorname{init}}^{\operatorname{f}}, p_{y,\operatorname{init}}^{\operatorname{f}}, v_{x,\operatorname{init}}^{\operatorname{f}}
    ]^\top \in \mathbb{R}^{10} ,
    \end{aligned}
    \end{equation}
 where 
$p_{x,\operatorname{init}}$, $p_{y,\operatorname{init}}$, $v_{x,\operatorname{init}}$, $p_{x,\operatorname{init}}^{\operatorname{c}}$, $p_{y,\operatorname{init}}^{\operatorname{c}}$, $v_{x,\operatorname{init}}^{\operatorname{c}}$, $p_{x,\operatorname{init}}^{\operatorname{f}}$, $p_{y,\operatorname{init}}^{\operatorname{f}}$, and $v_{x,\operatorname{init}}^{\operatorname{f}}$ represent the initial X-coordinate and Y-coordinate position as well as the longitudinal speed of the ego vehicle, dynamic chance and the nearest front vehicle, respectively. $\varphi_{\operatorname{init}}$ is the initial heading angle of the ego vehicle.

We further exploit a deep neural network to parameterize the policy $\pi$, with which the augmented decision vector $\mathbf{z}$ is modeled as
$
    \mathbf{z}=\pi\left(\mathbf{o}
    \right) = f_{\boldsymbol{\theta}}\left(\mathbf{o}
    \right),
$
where $\boldsymbol{\theta}$ are the parameters of the deep neural network, $\mathbf{o}$ is the observation vector of the vehicle. Moreover, 
we apply $z$-score normalization to input features in $\mathbf{o}$. In this work, we  present a novel CRL framework to determine the optimal policy $\pi^*$.

\subsection{Multi-Task Reward Formulation}

\textit{Sparse Lane-Change Reward}: The sparse reward function, which evaluates the quality of the optimal state trajectory $\boldsymbol{\xi}^*\left(\mathbf{z}\right)$ for the lane-change task, is designed as:

    \begin{equation}
    R_{\text{LC}}\left(\boldsymbol{\xi}^*\left(\mathbf{z}\right)\right)=R_{\max }- c_c \sum_{k=0}^N
    \rho_c\left|\boldsymbol{v}_k\right|^2,
    \label{reward_merging}
    \end{equation}   
where $\rho_c$ is a binary flag indicating whether a collision with surrounding vehicles occurs, $c_c \in \mathbb{R}_{+}$ is a hyperparameter for weighting the collision penalty, and $R_{\max } \in \mathbb{R}_{+}$ is the goal reward, which is gained when the ego vehicle merges into the traffic flow successfully. 
            
\textit{Reward Shaping:} 
The sparsity of the lane-change reward poses challenges to learning a viable policy within a reasonable time frame. Therefore, we introduce a novel reward term to directly evaluate the decision variables, which guides the RL agent to explore the policy space in directions of larger policy gradients:

\begin{align}
&R_{\text{DV}}\left(\mathbf{z}\right) = \nonumber\\
&-c_x \left|\Delta{p_x}\right| - c_y \rho_y \Delta{p_y} - c_t \left|t\right| 
- c_{\Delta t} \rho_{\Delta t} \Delta{t} 
 - c_{\Delta \varphi} \rho_{\Delta \varphi} \Delta \varphi \nonumber\\ 
& -  c_{p_x}\rho_{p_x} \left|{Q_{p_{x,\operatorname{tra}}}}\right|
-  c_{p_y}\rho_{p_y} \left|{Q_{p_{y,\operatorname{tra}}}}\right|
-  c_{\varphi}\rho_{\varphi} \left|{Q_{\varphi_{\operatorname{tra}}}}\right| \nonumber\\
& -  c_{v_x}\rho_{v_x} \left|{Q_{v_{x,\operatorname{tra}}}}\right|
-  c_{v_y}\rho_{v_y} \left|{Q_{v_{y,\operatorname{tra}}}}\right|
-  c_{\omega}\rho_{\omega} \left|{Q_{\omega_{\operatorname{tra}}}}\right|, \label{reward_dense}
\end{align}
where
$$
\begin{aligned}
&\Delta p_x  = p_{x, \operatorname{tra}} - p_{x,\operatorname{init}}^{\operatorname{c}},
\\ 
&\Delta p_y = \min(\left|p_{y, \operatorname{tra}} - p_{y, \operatorname{max}}\right|, \left|p_{y, \operatorname{tra}} - p_{y, \operatorname{min}}\right|),
\\
&\Delta t  = \min(\left|t_{\operatorname{tra}} - t_{\operatorname{max}} \right|, \left|t_{\operatorname{tra}} - t_{\operatorname{min}}\right|),
\\
&\Delta \varphi  = \min(\left|\varphi_{\operatorname{tra}} - \varphi_{\operatorname{max}}\right|, \left|\varphi_{\operatorname{tra}} - \varphi_{\operatorname{min}}\right|).
\\
\end{aligned}
$$   

In (\ref{reward_dense}), $t$ denotes the current time within the simulation episode,
$c_x$, $c_y$, $c_{t}$, $c_{\Delta t}$, $c_{\Delta \varphi}$, $c_{p_x}$, $c_{p_y}$, $c_{\varphi}$, $c_{v_x}$, $c_{v_y}$, and $c_{\omega}$ are weighting coefficients, $\rho_y$, $\rho_{\Delta t}$, and $\rho_{\Delta \varphi}$ are binary flags indicating whether the learnable position reference of Y-coordinate is 
within the range of values $[p_{y,\operatorname{min}},p_{y,\operatorname{max}}]$,
whether the reference of the learnable time  exceeds the simulation duration or falls below zero,
and whether the learnable heading angle exceeds the range of the feasible heading angle, respectively. $\rho_{p_x}$, $\rho_{p_y}$, $\rho_{\varphi}$, $\rho_{v_x}$, $\rho_{v_y}$, $\rho_{\omega}$ are also binary flags implying whether the learnable maximum of weighting is smaller than zero. 
$t_{\operatorname{min}}$ and $t_{\operatorname{max}}$ represent the lower bound and upper bound of the duration of simulation. $\varphi_{\operatorname{min}}$ and $\varphi_{\operatorname{max}}$ are the lower and upper bound of the predefined feasible heading angle range. 

\subsection{Curriculum Reinforcement Learning with Policy Transfer and Enhancement}

We operate the neural policy at the beginning of each episode to model $\mathbf{z}$, with which the MPC can generate a sequence of optimal trajectories $\boldsymbol{\xi}^*(\mathbf{z})$. Therefore, the corresponding reward signal $R(\boldsymbol{\xi}^*(\mathbf{z}))$ is received to evaluate the quality of generated trajectories. Hence, inspired  by \cite{wang2023learning}, the problem of finding the optimal neural policy with RL is reformulated to the following reward maximization problem:
\begin{equation}
\begin{aligned}
\centering
&\max _{\boldsymbol{\theta}} \quad R\left(\boldsymbol{\xi}^*\left(\mathbf{z}\left(\boldsymbol{\theta}\right)\right)\right) \\
&\quad \text{s.t. } \quad
\boldsymbol{\xi}^*\left(\mathbf{z}\left(\boldsymbol{\theta}\right)\right) = f_\mathrm{MPC}\left(\mathbf{z}\left(\boldsymbol{\theta}\right)\right).  
\end{aligned}
\end{equation}       
    
The gradient of the episode reward $R$ with respect to neural network parameters $\boldsymbol{\theta}$ is decomposed with chain rule, where
    \begin{equation}
     \frac{d R}{d \boldsymbol{\theta}}= \frac{\partial R}{\partial \mathbf{z}} \frac{\partial \mathbf{z}}{\partial \boldsymbol{\theta}}.
    \end{equation}  
    
Specifically, the sub-gradient $\boldsymbol{g_{s2}}=\partial \mathbf{z}  \verb|/| \partial  \boldsymbol{\theta}$ is calculated automatically by loss backward when updating the neural network. However, obtaining another sub-gradient $\boldsymbol{g_{s1}} = \partial R \verb|/| \partial \mathbf{z}$ is computationally expensive, as it requires differentiation through the nonlinear MPC optimization problem over the whole receding horizon. 

We denote the value in the $j$-th row in $\boldsymbol{g_{s1}}$ as $g_j$. With the finite difference policy gradient method, we can estimate $g_j$ as:
\begin{equation}
 g_j=\frac{R\left(\boldsymbol{\xi}^*\left(\mathbf{z}+\epsilon \boldsymbol{e_j}\right)\right)-R\left(\boldsymbol{\xi}^*\left(\mathbf{z}\right)\right)}{\epsilon},
 \label{finit_diff}
\end{equation}
where $\boldsymbol{e_j}$
is a unit vector with 1 in the $j$-th row, $\epsilon$ is a small random step in the direction $\boldsymbol{e_j}$ to perturb the augmented decision vector $\mathbf{z}$. 

We exploit on-policy RL to train the neural
network policy through the maximization of the designed reward.
Nevertheless, the reward in such hierarchical framework does not exhibit a direct correspondence with the neural network output. That is to say, unfavorable decision variables can still yield an acceptable reward due to the inherent robustness of MPC.
Furthermore, due to the sparsity of lane-change reward, the sufficiency of policy exploration is not guaranteed, so it is challenging to yield an acceptable policy in finite time. 
Consequently, the training of neural network suffers from inefficient and insufficient exploration in the policy space and even unstable learning.

Curriculum learning (CL) is a well-established routine to accelerate the exploration when RL is handling rather complex missions, especially addressing extrapolation error and premature policy convergence \cite{song2021autonomous}.
Therefore, we present a CRL framework to train our neural policy with policy transfer and policy enhancement. We generate three modes of curricula, which are represented by $\mathscr{C}=\left\{C_i\right\}, i\in\left\{1,2,3\right\}$. The curricula are designed with different tasks in different domains.


\textit{Curriculum 1: Transferable Policy Learning with Reward Shaping in Static Environment.}
In source domain 1 which is denoted as $\mathcal{M}_{s,1}$, the traffic flow on the fast lane keeps static. The objective of Curriculum 1 is to learn a transferable neural policy. We train our randomly initialized neural network through the maximization of the reward term directly evaluating the decision variables as \eqref{reward_dense}.
In $\mathcal{M}_{s,1}$, the task $\mathcal{T}_{s,1}$ is to train an acceptable neural policy $\pi_{s,1}$ in finite time, with which the learnable state references $\mathbf{x}_\mathrm{tra}$, maximum of weighting matrix $\mathbf{Q}_{\max}$, and tracking time reference $t_\mathrm{tra}$ are expected to converge to a feasible range roughly. Therefore, the RL agent is guided by the empirically-designed reward, increasing the exploration efficiency in the policy space.

\textit{Curriculum 2: Lane-Change Policy Learning Under Low-Speed Setting.}
In Curriculum 2, we load the transferable policy $\pi_{s,1}$ and train it in source domain $\mathcal{M}_{s,2}$, where the traffic flow on the fast lane move at a speed lower than the normal setting. The objective of Curriculum 2 is to generalize the transferable $\pi_{s,1}$ to a lane-change policy $\pi_{s,2}$ by maximizing the lane-change reward as (\ref{reward_merging}).

\textit{Curriculum 3: Lane-Change Policy Enhancement Under Normal-Speed Setting.}
In Curriculum 3, we aim to obtain an optimal neural policy $\pi^*$ for the lane-change task under a normal speed setting. 
In the target domain $\mathcal{M}_{t}$, the traffic flow on the fast lane moves at a normal speed and the weight of the penalty term in terms of collisions in lane-change reward is increased. We load and enhance the lane-change policy $\pi_{s,2}$, and obtain the optimal policy $\pi^*$ eventually by maximizing the revised lane-change reward.

To this end, the proposed CRL framework with policy transfer and enhancement is summarized in Algorithm \ref{algo-learning}.

\begin{algorithm}[t]  
	\caption{Curriculum Reinforcement Learning with Policy Transfer and Enhancement} \label{algo-learning}
	\LinesNumbered 
	\KwIn{$f_\mathrm{MPC}$, $\mathscr{C}$}
	\KwOut{$\pi^* = f_{\boldsymbol{\theta}^*}$}
        Initialize $\boldsymbol{\theta}$\; 
        \While{not terminated}  
            {Select Curriculum $C_i$ from Curricula $\mathscr{C}$\;
		  Reset the environment to get $\mathbf{o}$ and $\mathbf{x}_g$ according to $C_i$\;
            \If{curriculum switched}
                {Load policy $\pi^*$ trained with $C_{i-1}$\;}
            Compute $\mathbf{z}=f_{\boldsymbol{\theta}}(\mathbf{o})$\;
            Solve MPC($\mathbf{z}$) as (\ref{MPC}) online to obtain $\boldsymbol{\xi}^*\left(\mathbf{z}\right)$ \;
            \For{$j \leftarrow 1$ \KwTo $dim(\mathbf{z})$}
		  {Perturb $j$-th row in $\mathbf{z}$\;
                Estimate $g_j$ using (\ref{finit_diff})\;}
            Update $\boldsymbol{\theta}$ using gradient ascent with $\boldsymbol{g_{s1}}$ and $\boldsymbol{g_{s2}}$\;}   
\end{algorithm}

\section{Experiments}
\label{sec:exp}

\subsection{Numerical Setup}

The MPC problem is solved using CasADi \cite{andersson2019casadi} with IPOPT. In MPC, we set the receding horizon and the step time to $T = \SI{5.0}{s}$ and $d_t = \SI{0.1}{s}$. Furthermore, we take the weighting matrices $\mathbf{Q}_{x}$, $\mathbf{Q}_{u}$, and $\mathbf{Q}_{\Delta_u}$ to $\operatorname{diag}\left(\left[100, 100, 100, 10\right]\right)$, $\operatorname{diag}\left(\left[1, 1\right]\right)$, and $\operatorname{diag}\left(\left[0.1, 0.1\right]\right)$, respectively. Also, we set the lower and upper bounds of acceleration and steering angle to $a_{\min} = \SI{-6.0}{m/s^2}$, $a_{\max} = \SI{3}{m/s^2}$, $\delta_{\min} = \SI{-0.6}{rad}$, and $\delta_{\max} = \SI{0.6}{rad} $, respectively. 

The deep neural network is constructed in PyTorch \cite{paszke2019pytorch}, with the structure of 4 hidden layers with 128 LeakyReLU nodes. The neural network is trained by Adam optimizer \cite{kingma2014adam} with an initial learning rate $3 \times 10^{-4}$, where the learning rate decays in 0.96 every 32 steps. Weights and Biases \cite{wandb} is utilized to monitor the training process.

We train the policy network and evaluate the driving performance in numerical simulations, where the ego vehicle is placed at $\left[ p_{x,\operatorname{init}} \sim \mathcal{N}(30, 2.5), -2.5\right]$ and the dynamic chance moves from $\left[ p_{x,\operatorname{init}}^{\operatorname{c}} \sim \mathcal{N}(50, 10), 2.5\right]$ at a time-variant speed of $v_{x}^{\operatorname{c}} \sim \mathcal{N}(\mu_i, \sigma_i)$ $\SI{}{m/s}$. Here, $\mu_i$ and $\sigma_i$ are the mean speed and speed standard deviation of each curriculum from $i \in \left\{1,2,3\right\}$. In our settings, $\mu_1, \sigma_1, \mu_2, \sigma_2, \mu_3, \sigma_3$ are $0, 0, 2, 0.5, 4, 1$, respectively. The goal states $\mathbf{x}_g$ are set to 
$
\left[\begin{array}{llllll}
    p_{x}^{\operatorname{c}}& 2.5& 0& v_{x}^{\operatorname{c}}& 0& 0
    \end{array}\right]^{\top},
$ 
where $p_{x}^{\operatorname{c}}$ is the X-coordinate position of the dynamic chance obtained from traffic in real time.
 
\subsection{Training Result}

In order to illustrate the effectiveness of the proposed method (denoted as MPC-CRL) in policy learning, we employ the subsequent learning-based baselines for comparison:
\begin{itemize}
    \item High-level MPC with augmented decision variables and vanilla RL (denoted as MPC-RL).
    \item High-level MPC with SE(3) decision variables \cite{wang2023learning} and CRL (denoted as MPC-SE3-CRL).
\end{itemize}

\begin{figure}[t] 
    \centering   
    \includegraphics[trim=0 0 0 0, width=0.9\linewidth]{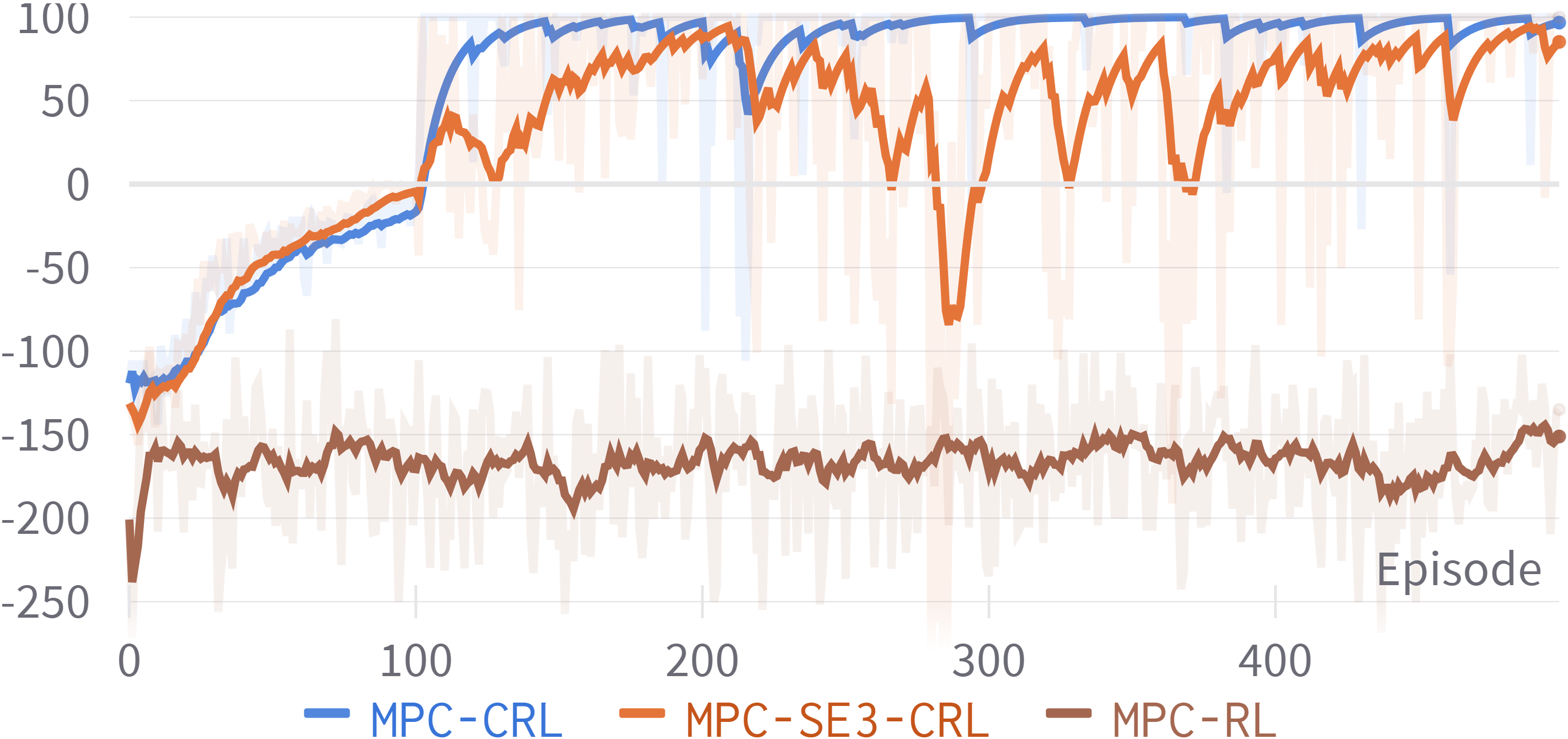}
   \caption{Reward curves of different methods. The training curves are smoothed by exponential moving average with a degree of 0.8, and the curriculum is switched at episodes 100 and 200.}
    \label{training_results}
\end{figure}

The reward curves are presented in Fig.~\ref{training_results}. The results exhibit that our proposed MPC-CRL surpasses MPC-RL both in terms of convergence speed and reward performance. 
Additionally, the incorporation of augmented decision variables, which adaptively modulate the costs of MPC, endows MPC-CRL with superior reward performance when compared to MPC-SE3-CRL.
Therefore, the training results clearly indicate that the presented CRL framework effectively encourages the RL agent to efficiently and sufficiently explore the policy space, ultimately leading to the attainment of the satisfactory optimum. Additionally, the introduction of augmented decision variables notably improves the lane-change performance.

\subsection{Performance Evaluation}

\begin{figure}[t] 
    \centering   
    \includegraphics[trim=1cm 0 0 0, width=0.9\linewidth]{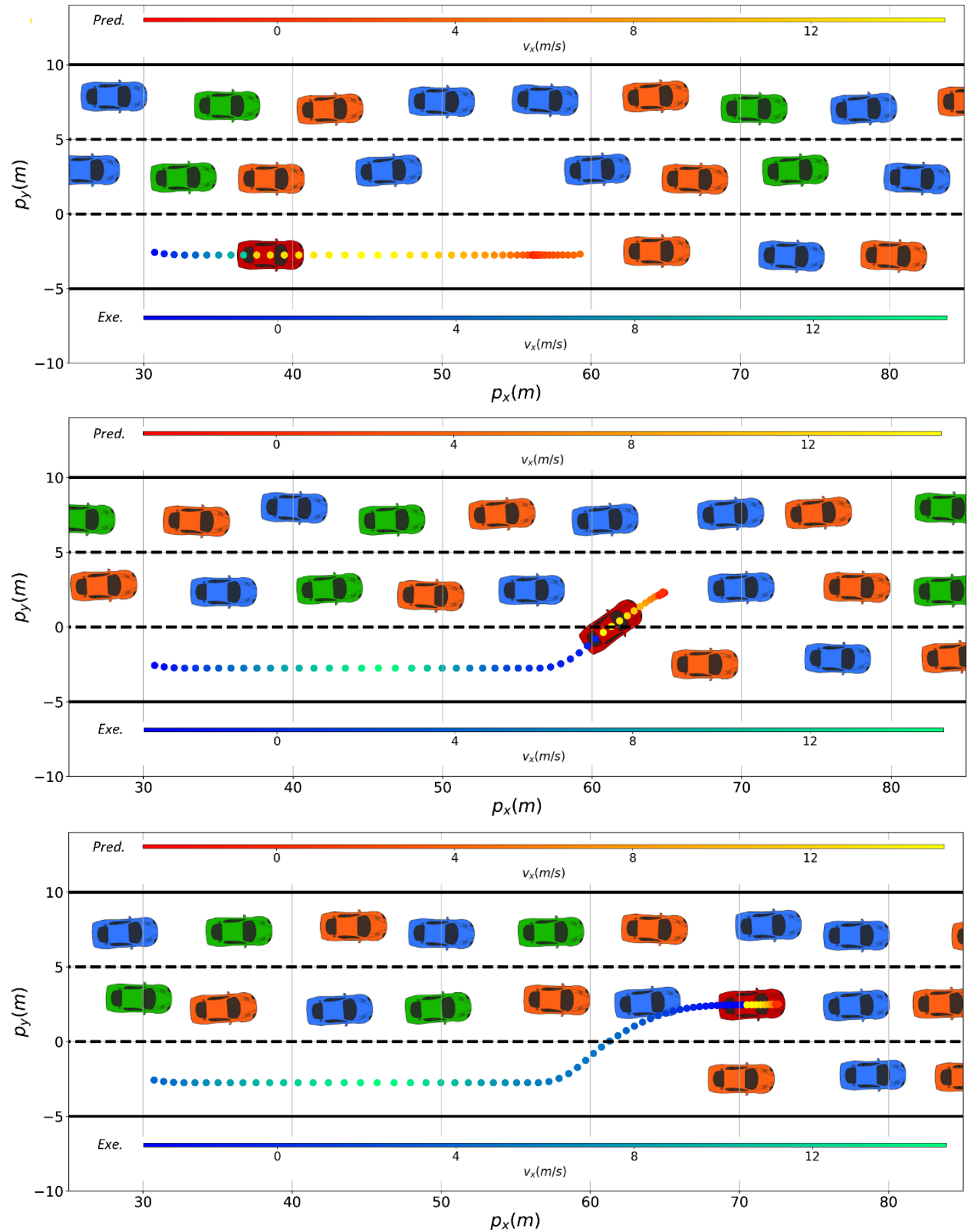}
   \caption{Key frames of a trail with our proposed approach for chance-aware lane change in numerical simulations. The vehicles on the middle and upper lane represent the traffic flow, the vehicle in red is the ego vehicle, the vehicles ahead of the ego vehicle on the lower lane are front vehicles, the dotted line in red and blue represent the future trajectory and the executed trajectory of the ego vehicle. The colorbars refer to the different longitudinal speed of predicted and executed trajectories of the ego vehicle.}
    \label{simu}
\end{figure}

\begin{figure}[t] 
    \centering   
    \includegraphics[trim=0 0 0 0, width=0.9\linewidth]{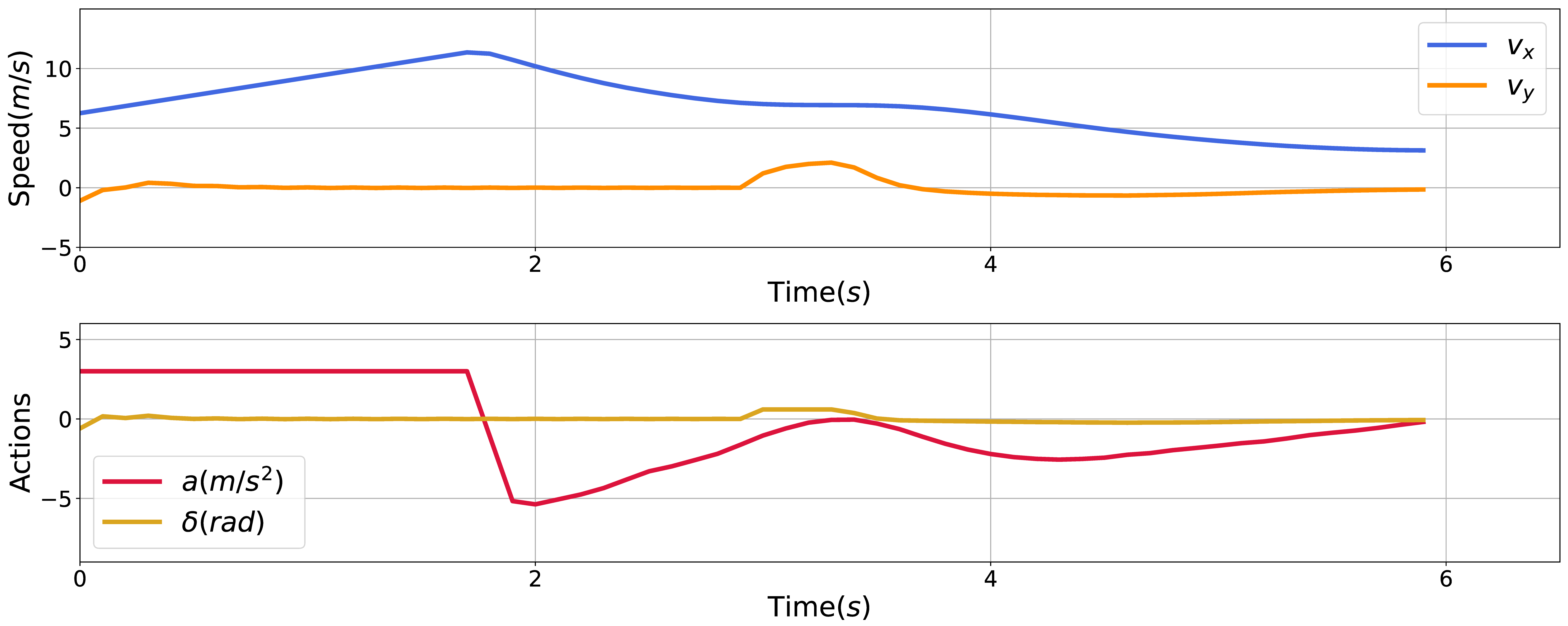}
   \caption{The speed and action profiles of a trail of chance-aware lane change. }
    \label{value}
    \vspace{-0.2cm}
  \end{figure}

To provide further insight into the driving performance attained by our framework, a set of trails with various settings are conducted for driving performance evaluation, where a series of snapshots from a trail are presented in Fig.~\ref{simu}, and the corresponding speed and action profiles are illustrated in Fig.~\ref{value}.
As shown in Fig.~\ref{simu}(a), when $t=\SI{1.0}{s}$, the ego vehicle plans a long trajectory along the lower lane, intending to track the learnable state references $\mathbf{x}_\mathrm{tra}$. Simultaneously, the ego vehicle accelerates and exploits the maximum of bounded acceleration, as indicated in Fig.~\ref{value}. Then, Fig.~\ref{simu}(b) visualizes the ego vehicle's subsequent actions, wherein it decelerates and executes a left turn, preparing of a transition to the middle lane with enough safe margin at $t=\SI{3.8}{s}$. The recorded longitudinal speed at this moment is approximately $\SI{6.1}{m/s}$. Ultimately, as depicted in Fig.~\ref{simu}(c), the ego vehicle successfully occupies the recognized dynamic chance and safely merges into the traffic flow on the middle lane at $t=\SI{5.9}{s}$.

\vspace{0.2cm}

\subsection{Comparison Analysis}
Due to the failure of acquiring essential domain knowledge during training, MPC-RL does not qualify as a suitable baseline for the comparison of driving performance.
We further exploit a new baseline, which adopts the paradigm of high-level MPC with augmented decision variables curated through human-expert experience (denoted as MPC-HE). Moreover, to quantitively analyze the results attained by different methods, 
{we define a collision-free episode finished within finite time as a successful case. Otherwise, we record an episode involving any collision as a collided case and an episode terminated by the episode duration of simulation as a time-out case.}
Afterwards, we run the trails repeatedly 100 times and record the corresponding success, collision and time-out rate of various methods, as documented in Table \ref{results_tab}. 
The results indicate that our proposed approach outperforms all baselines in terms of both task accomplishment and safety assurance due to the highest success rate of $96\%$ and the lowest collision rate of $4\%$. By leveraging augmented decision variables to automatically modulate the costs of MPC, the adaptiveness of our approach to dense and dynamic traffic is improved significantly. Hence, we conclude that our proposed framework manifests a stronger guarantee of collision avoidance and task success.

\begin{figure}[t] 
    \centering   
    \includegraphics[trim=0 0 0 0, width=1.0\linewidth]{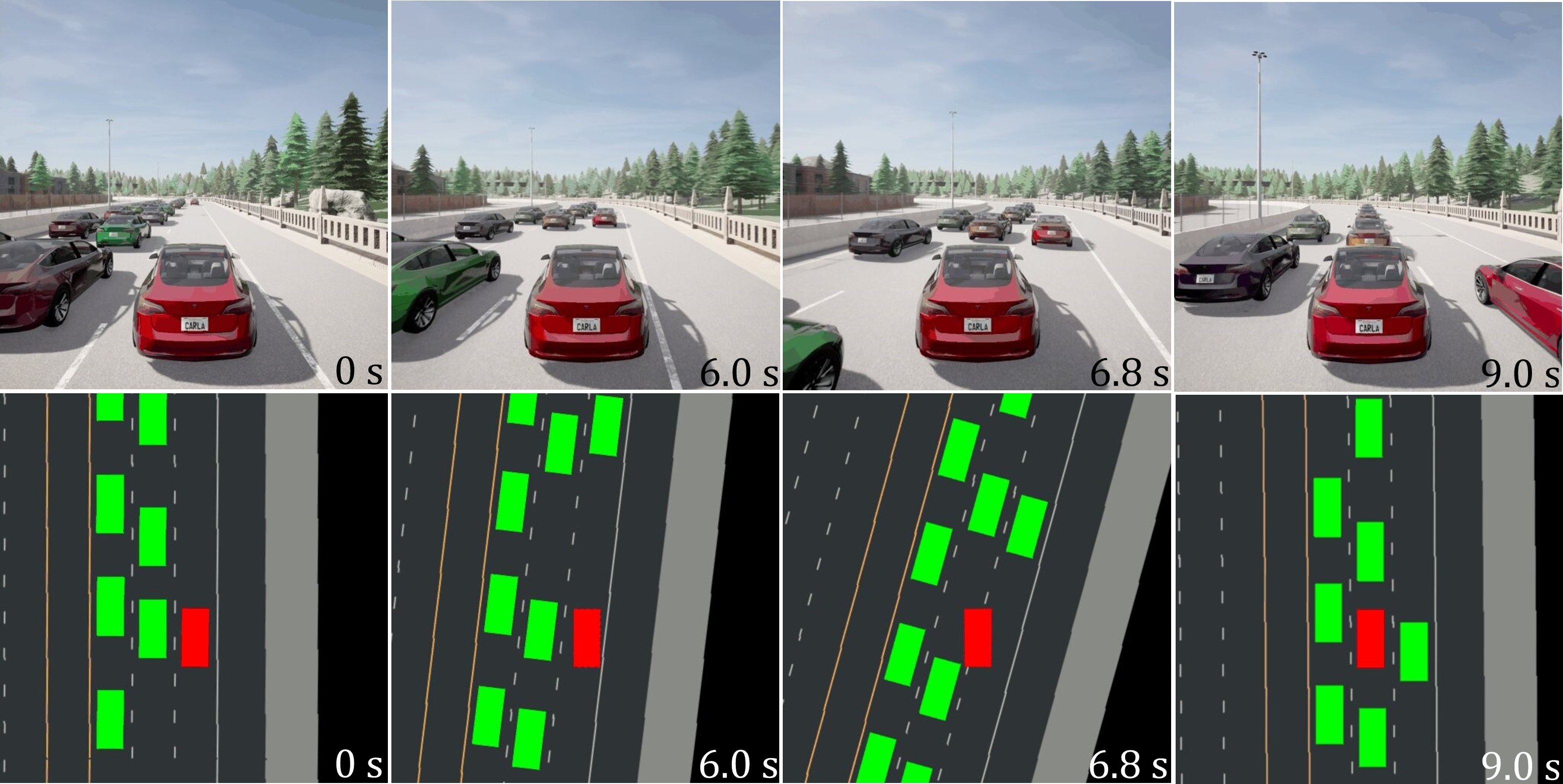}
   \caption{Key frames of the experimental validation of our method in the high-fidelity simulator. The top shows the third-person view attached to the ego vehicle. The bottom shows the bird-eye view, where the red rectangle is the ego vehicle while the green rectangles denote the surrounding vehicles. }
    \label{validation}
    \vspace{-0.2cm}
\end{figure}

\begin{table}[t]
    \centering
    \caption{Success, collision, and time-out rate of different methods for chance-aware lane-change tasks.}\label{results_tab}
    \begin{tabular}{cccc}
        \toprule
        Approaches & Succ. (\%) & Coll. (\%) & Time-out (\%) \\
        \midrule
        \textbf{MPC-CRL}  & $\bm{96}$ & $\bm{4}$  & $\bm{0}$ \\
            MPC-SE3-CRL    & 77 & 23  & 0 \\ 
            MPC-HE   & 69 & 28 & 3\\
       
        \bottomrule
    \end{tabular}
\end{table}

\subsection{Validation in High-Fidelity Simulator}

We further validate the effectiveness of our proposed framework in the high-fidelity simulator CARLA \cite{dosovitskiy2017carla}. The outer ring road with three lanes in Town05 is the testbed for performance validation, where all traffic participants are set to Tesla Model~3. In order to reduce the gap between the environments in numerical simulations and high-fidelity validations, the global Cartesian coordinate frame is transformed to the coordinate frame of centerline reference \cite{eiras2021two}. We load the policy model trained in numerical simulations and fine-tune it. Key frames of a representative example are presented in Fig. \ref{validation}. 
The results demonstrate clearly that the RL agent trained by our approach retains the capability to effectively determine suitable maneuvers and appropriate timing for a lane change in CARLA. 
The validation results highlight the efficacy and generalizability of our method when deployed to the high-fidelity simulator.

\section{Conclusions}
\label{sec:conc}
In this paper, we proposed a novel learning-based MPC framework with appropriate parameterization using augmented decision variables. 
Instead of choosing partial variables (such as only the positions) as references, we utilized a neural policy to learn full-state references and regulatory factors corresponding to their relative importance. 
Hence, the cost terms were automatically modulated with our specifically-designed augmented decision variables.
Furthermore, ordered multi-phase curricula were generated for learning a neural policy using RL, which leads to faster convergence speed and better policy quality.
Furthermore, through a series of comparative experiments, our approach demonstrated superiority in terms of success rate in chance-aware lane-change tasks under dense and dynamic traffic settings. 
Moreover, the practicality and generalizability of our method \textcolor{black}{were} further illustrated through the experimental validations in the high-fidelity simulator.




\bibliographystyle{IEEEtran}
\bibliography{ref.bib}

\end{document}